\documentclass[a4paper,twoside, french]{article}
\usepackage{babel}
\usepackage{rfia2000}
\usepackage{times}
\usepackage{graphicx}
\usepackage{amsmath}
\usepackage{xcolor}
\usepackage{float}
\usepackage[french, onelanguage, ruled, noend]{algorithm2e}
\usepackage[noend]{algpseudocode}
\usepackage[T1]{fontenc}
\usepackage[utf8]{inputenc}
\usepackage{bm}
\usepackage{caption}
\usepackage{subcaption}
\usepackage[hidelinks]{hyperref}
\usepackage{parskip}
\usepackage{placeins}
\usepackage{stfloats}
\usepackage{enumitem}
\SetAlFnt{\scriptsize}
\let\Oldsection\section
\renewcommand{\section}{\FloatBarrier\Oldsection}
\let\Oldsubsection\subsection
\renewcommand{\subsection}{\FloatBarrier\Oldsubsection}
\let\Oldsubsubsection\subsubsection
\renewcommand{\subsubsection}{\FloatBarrier\Oldsubsubsection}

\SetCommentSty{mycommfont}
\usepackage[compress,numbers]{natbib}
\begin{document}
%%%%%Pas de date
\date{}
%%%%% Titre gras 14 points
\title{\Large\bf TSRuleGrowth: Extraction de règles de prédiction semi-ordonnées à partir d'une série temporelle d'éléments discrets, application dans un contexte d'intelligence ambiante}
\author{
	\begin{tabular}[t]{c@{\extracolsep{.5em}}c@{\extracolsep{.5em}}c@{\extracolsep{.5em}}c@{\extracolsep{.5em}}c}
		Benoit Vuillemin${}^{1,2}$ & Lionel Delphin-Poulat${}^2$ & Rozenn Nicol${}^2$ & Laetitia Matignon${}^1$ & Salima Hassas${}^1$\\
	\end{tabular}
	{} \\
	\\
	${}^1$        Univ Lyon, Université Lyon 1, CNRS, LIRIS, UMR5205, F-69622, France   \\
	${}^2$        Orange Labs, Lannion, France
	{} \\
	\\
	\{benoit.vuillemin, laetitia.matignon, salima.hassas\}@liris.cnrs.fr\\ \{lionel.delphinpoulat, rozenn.nicol\}@orange.com\\
}
\maketitle
%%%%  Pas de num\'erotation sur la page de titre
\thispagestyle{empty}
\subsection*{R\'esum\'e}
{\em
Cet article présente un nouvel algorithme : TSRuleGrowth, recherchant des règles de prédiction semi-ordonnées sur une série temporelle. Cet algorithme reprend les principes de l'état de l'art de la fouille de règles et les applique aux séries temporelles via une nouvelle notion de support. Nous l'appliquons à des données réelles provenant d'un environnement connecté. Cet algorithme extrait les habitudes des utilisateurs à travers différents objets connectés.
}
\subsection*{Mots Clef}
Fouille de règles, Intelligence Ambiante, Habitudes, Automatisation, Support, Séries temporelles

\subsection*{Abstract}
{\em
This paper presents a new algorithm: TSRuleGrowth, looking for partially-ordered rules over a time series. This algorithm takes principles from the state of the art of rule mining and applies them to time series via a new notion of support. We apply this algorithm to real data from a connected environment, which extract user habits through different connected objects.
}
\subsection*{Keywords}
Rule Mining, Ambient Intelligence, Habits, Automation, Support, Time series

\section{Introduction} \label{search-introduction}
La fouille de règles de prédiction sur une série temporelle est un problème majeur en data mining. Utilisé notamment dans l'analyse du cours des actions et la recommandation d'achats, ce problème est de plus en plus étudié à mesure que le domaine de l'intelligence ambiante (AmI) se développe. L'AmI est la fusion entre l'intelligence artificielle et l'internet des objets, et peut être décrit comme : ``Un environnement numérique qui soutient les personnes dans leur vie quotidienne de façon proactive, mais raisonnable'' \cite{augustoAmbientIntelligenceConcepts2007}. Ce travail entre dans le domaine de l'AmI : nous voulons faire un système qui trouve les habitudes des utilisateurs dans un environnement connecté, autrement dit un environnement dans lequel des objets connectés sont présents, afin de fournir aux utilisateurs des automatisations. Par exemple, si une personne allume habituellement la lumière après être entrée chez elle, ce qui peut être vu par une ampoule connectée et un capteur de porte, le système pourrait détecter cette règle de prédiction et proposer d'allumer la lumière à chaque entrée de l'utilisateur.\\
Cet article décrit un nouvel algorithme utilisé dans notre système d'AmI, qui vise à rechercher les règles de prédiction sur une série temporelle. Ici, la série temporelle représente les événements envoyés par les objets connectés. Ces règles de prédiction seront ensuite proposées aux utilisateurs comme suggestions d'automatisation. Dans le cadre de cet article, les séries temporelles sont composées uniquement de valeurs catégorielles, plutôt que continues. Les données peuvent survenir à tout moment, c'est-à-dire qu'il n'y a pas de fréquence d'échantillonnage fixe dans les séries temporelles. La structure de ces règles est expliquée, ainsi que l'état de l'art des domaines concernés, ce qui justifie les choix effectués pour cet algorithme.
\section{Contexte et définitions}\label{search-motivation}
\subsection{Entrée}\label{search-motivation-input}
\begin{figure}
	\centering
	\includegraphics[width=\linewidth]{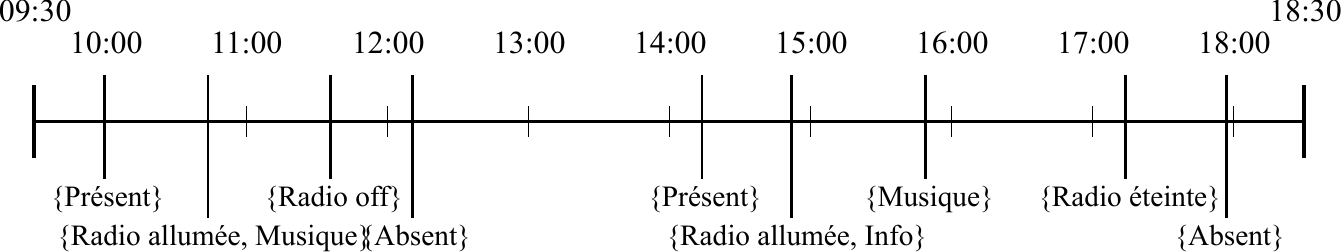}
	\caption{Représentation d'une série temporelle}
	\label{fig:TimeSeries}
\end{figure}

Il existe deux types d'objets connectés : les capteurs et les actionneurs. Les \textbf{capteurs} observent des grandeurs physiques de l'environnement. Un capteur renvoie des événements, correspondant aux changements d'état de la variable observée. Un capteur d'ouverture de porte, par exemple, peut renvoyer des événements d'ouverture et de fermeture. Les capteurs peuvent mesurer des variables \textbf{continues} ou \textbf{catégorielles}. Par exemple, la température d'une pièce, exprimée en degrés, peut être considérée comme variable continue, tandis que la sélection d'une station radio ou l'ouverture d'une porte sont des variables catégorielles. \textbf{Dans cette étude, seuls les capteurs renvoyant des variables catégorielles sont considérés.} Les \textbf{actionneurs} agissent sur l'environnement. Un actionneur renvoie un événement lorsqu'il a effectué une action. Par exemple, un volet connecté renvoie un événement lorsqu'il s'ouvre ou se ferme. De la même manière que les capteurs, les actionneurs peuvent effectuer des actions catégorielles, comme ouvrir un volet, ou continues, comme augmenter la température à une certaine valeur. \textbf{Comme pour les capteurs, seuls les actionneurs qui effectuent des actions catégorielles sont pris en compte.}\\
Chaque objet, qu'il soit un capteur ou un actionneur, renvoie des événements primaires. Dans cet article, ils sont nommés \textbf{éléments}. Tous les éléments envoyés par tous les objets sont regroupés dans un ensemble noté $E$.\\
Prenons l'exemple d'une pièce contenant deux objets connectés : un détecteur de présence, permettant de savoir si une personne se trouve dans une pièce ou non, et un actionneur : une radio. Le détecteur de présence peut renvoyer les éléments suivants : ``Présent'' et ``Absent''. La radio peut agir de deux façons : son état peut être ``Radio allumée'' or ``Radio éteinte'', et elle peut sélectionner l'une des stations suivantes : ``Musique'', ``Info'', ``Débat''. Ainsi, l'ensemble de tous les éléments est $E$ = \{Présent, Absent, Radio allumée, Radio éteinte, Musique, Info, Débat\}.\\
Le système AmI que nous proposons recueille des flux de données à partir de plusieurs objets connectés. Chaque flux de données est composé d'une succession d'éléments, dont chacun peut se produire une ou plusieurs fois. Chaque occurrence est estampillée d'un timestamp, une donnée temporelle. Ainsi, chaque élément est potentiellement associé à plusieurs timestamps correspondant à ses multiples occurrences. Pour la suite du traitement, toutes les données collectées par les différents capteurs sont regroupées en une seule \textbf{série temporelle}. En d'autres termes, une série temporelle est obtenue par une juxtaposition dans le temps d'éléments fournis par tous les objets. Elle est notée $TS$=$\langle(t_1, I_1), ..., (t_{n}, I_{n})\rangle, I_1, ..., I_{n} \subseteq E$, où:
\begin{itemize}
\item $t_i$ est un \textbf{timestamp}, un point fixe dans le temps.
\item $I_i \subseteq E$ est un \textbf{itemset}. C'est l'ensemble des éléments uniques provenant de $E$ observés au timestamp $t_i$.
\end{itemize}
Il est à noter qu'un élément ne peut être vu qu'une seule fois dans un itemset. De plus, les timestamps ne sont pas nécessairement espacés de de manière uniforme. La figure \ref{fig:TimeSeries} est un exemple de série temporelle créée à partir de l'environnement connecté mentionné dans la section \ref{search-motivation-input}.
Sa représentation mathématique est :\\
$TS=\langle$(10:00 am, \{Présent\}), (10:44 am, \{Radio allumée, Musique\}), (11:36am, \{Radio éteinte\}), (12:11 am, \{Absent\}), (2:14 pm, \{Présent\}), (2:52 pm, \{Radio allumée, Informations\}), (3:49 pm, \{Musique\}), (5:14 pm, \{Radio éteinte\}), (5:57 pm, \{Absent\})$\rangle$.\\
Cette série temporelle représente certaines actions d'un utilisateur dans l'environnement. La section suivante détaille ce que le système doit trouver dans une série temporelle.
\subsection{Sortie}\label{search-motivation-output}
Le système proposé doit trouver des \textbf{règles de prédiction}, pour exprimer les habitudes observées. Une règle de prédiction est notée $R:E_{c} \Rightarrow E_{p}$, où $E_{c}$ est la \textbf{condition}, et $E_{p}$ est la \textbf{prédiction} de la règle. $R$ décrit que si $E_{c}$ est observé, $E_{p}$ sera observé après un certain temps.\\
Dans le cas d'utilisation étudié, nous voulons limiter la recherche de règles \textbf{pour lesquelles la partie prédiction $E_{p}$ doit être composée uniquement d'éléments provenant d'actionneurs}. En effet, la recherche de règles étant fortement combinatoire, cela permet de limiter cet aspect tout en s'adaptant au cas d'utilisation : proposer des automatisations d'actions en fonction de situations. D'après la série temporelle de la section \ref{search-motivation-input}, une règle peut être $\{\text{Présent}\} \Rightarrow \{\text{Radio allumée}\}$. Il s'agit d'une règle basique, où la condition et la prédiction ne sont composés que d'un seul élément. Nous ne voulons pas, par exemple, trouver la règle $\{\text{Radio éteinte}\} \Rightarrow \{\text{Absent}\}$ car la partie prédiction, $\{\text{Absent}\}$, provient d'un capteur (de présence) et non d'actionneurs (comme la radio).\\
Une règle, pour être validée, doit être fréquente et fiable. Il est aussi possible de faire de la détection d'anomalies, c'est-à-dire rechercher des règles très peu fréquentes et très intéressantes, mais cela n'entre pas dans le cadre de cet article.\\
Plusieurs types de règles de prédiction sont possibles \cite{fournier-vigerERMinerSequentialRule2014}:
\begin{itemize}
	\item \textbf{Règles séquentielles complètement ordonnées}, où la condition $E_{c}$ et la prédiction $E_{p}$ sont des séquences, c'est-à-dire des successions temporelles d'éléments,
	\item \textbf{Règles séquentielles semi-ordonnées} \cite{fournier-vigerMiningPartiallyOrderedSequential2015}, où la condition $E_{c}$ et la prédiction $E_{p}$ ne sont pas ordonnées. Un ordre existe toujours entre la condition et la prédiction, d'où le nom ``semi-ordonné''. Deux structures mathématiques sont possibles pour $E_{c}$ et $E_{p}$ : les \textbf{ensembles}, où un élément ne peut apparaître qu'une fois, et les \textbf{multiensembles}, où plusieurs instances d'éléments sont possibles. Le nombre d'instances d'un élément dans un multiensemble est appelé \textbf{multiplicité}. Par exemple, la multiplicité de l'élément $x$ dans le multiensemble $\{x, x, y\}$ est 2.
\end{itemize}
Après avoir testé chacun des types, nous avons choisi d'utiliser des \textbf{règles séquentielles semi-ordonnées contenant des multiensembles}. Le problème avec les règles séquentielles complètement ordonnées est que plusieurs règles peuvent caractériser la même situation. Par définition, l'extraction de règles semi-ordonnées génère moins de candidats et moins de règles. De plus, elles sont décrites comme étant plus générales, avec une plus grande précision de prédiction que l'autre type de règles, et elles ont été utilisées dans des applications réelles \cite{fournier-vigerERMinerSequentialRule2014}. Dans le cas d'utilisation proposé, la description d'une situation ne nécessite pas nécessairement un ordre, mais la multiplicité d'un élément peut être significative. Pour expliquer ce choix, prenons l'exemple d'une lampe à détection sonore. Lorsque l'on tape deux fois dans les mains, que l'on fait deux fois le même son, la lampe s'allume.
\section{Travaux associés}\label{search-relatedwork}
Comme dit ci-dessus, le système proposé doit rechercher des règles de prédiction semi-ordonnées sur une série temporelle d'éléments provenant de capteurs et d'actionneurs. Ainsi, dans l'état de l'art, deux grands domaines de recherche doivent être considérés : la fouille de règles sur les séries temporelles et la fouille de règles semi-ordonnées. Mais avant cela, rappelons quelques définitions.\\
Une règle de prédiction $R: E_{c} \Rightarrow E_{p}$ doit être fréquente et fiable. Pour vérifier qu'une règle est fréquente, son \textbf{support} est calculé. La notion de support dépend de la structure des données en entrée, mais il estime la fréquence d'une règle, d'un ensemble d'éléments ou d'un élément. Pour vérifier qu'une règle est fiable, son \textbf{intérêt} est calculé. Plusieurs mesures permettent de connaître l'intérêt d'une règle. La plus connue est la confiance \cite{azevedoComparingRuleMeasures2007}, mais il existe des alternatives, telles que la conviction, le lift \cite{azevedoComparingRuleMeasures2007} ou netconf \cite{febrer-hernandezSPaCNFClassifierBased2016, ahnEfficientMiningFrequent2004}. Ces mesures dépendent des supports de $R$, $E_{c}$ et $E_{p}$.
\subsection{Fouille de règles sur séries temporelles}\label{search-rulemining}
\cite{dasRuleDiscoveryTime1998} propose un système de fouille de règles basiques, où un élément en prédit un autre, sur une séquence d'éléments. Ces éléments représentent des variations basiques de données boursières. Il peut aussi rechercher des règles plus complexes, où la condition est une séquence. Ce système permet donc de trouver des règles sur une série temporelle. Cependant, il cherche des règles complètement ordonnées, plutôt que des règles semi-ordonnées. De plus, la partie prédiction des règles est limitée à un seul élément, une limitation que nous voulons éviter dans ce système d'AmI.\\
\cite{schluterAnalysisTimeSeries2011} peut être considéré comme une amélioration de \cite{dasRuleDiscoveryTime1998}, car il recherche des règles où la prédiction n'est pas limitée à un élément. Mais, recherchant des règles complètement ordonnées, il ne peut pas être appliqué dans notre cas.\\
\cite{mannilaDiscoveryFrequentEpisodes1997} introduit une notion de support pour série temporelle, via une fenêtre glissante à durée fixe. Le support d'un élément, d'un ensemble d'éléments ou d'une règle est le nombre de fenêtres dans lesquelles cet élément, ensemble ou règle apparaît. L'algorithme trouve des règles semi-ordonnées, en cherchant des ensembles fréquents d'éléments, puis en les combinant pour générer des règles. D'autres algorithmes utilisent ce support, dont \cite{deogunPredictionMiningApproach2005} qui trouve des règles dont la prédiction est composée d'un élément.\\
L'algorithme présenté dans \cite{mannilaDiscoveryFrequentEpisodes1997} peut donc s'appliquer dans notre cas. Cependant, la définition du support peut être problématique. En effet, les éléments $E_{p}$ étant strictement postérieurs à ceux de $E_{c}$, le nombre de fenêtres recouvrant la règle $R$ est strictement inférieur à celui de $E_{c}$. Ainsi, même si $E_{p}$ apparaît toujours après $E_{c}$ dans un temps donné, le support de la règle est inférieur à celui de $E_{c}$, réduisant son intérêt. De plus, comme la recherche est structurée en deux étapes (recherche d'ensembles fréquents, puis recherche de règles), l'algorithme n'est pas totalement efficace.
\subsection{Fouille de règles semi-ordonnées}\label{search-semisequential}
A notre connaissance, peu d'algorithmes de fouille de règles semi-ordonnées existent. Les plus connues sont RuleGrowth \cite{fournier-vigerMiningPartiallyOrderedSequential2015}, et ses variations, TRuleGrowth \cite{fournier-vigerMiningPartiallyOrderedSequential2015} et ERMiner \cite{fournier-vigerERMinerSequentialRule2014}. Ces algorithmes prennent en entrée un ensemble de transactions. Une transaction est une séquence d'ensembles d'éléments appelés itemsets, ordonnée dans le temps, mais contrairement aux séries temporelles, sans timestamp associé.
Pour vérifier qu'une règle est fréquente, ces algorithmes calculent son support, lié à la structure des transactions. Pour vérifier qu'une règle est fiable, ils calculent son intérêt, lié au support de la règle, de la condition et de la prédiction.
TRuleGrowth est une extension de RuleGrowth qui accepte la contrainte d'une fenêtre glissante, définie comme un nombre d'itemsets consécutifs. Il permet de limiter la recherche aux règles qui ne peuvent se produire que dans cette fenêtre. ERMiner est une version plus efficace de RuleGrowth, mais sans fenêtre glissante.\\
Ces algorithmes cherchent directement des règles de prédiction, contrairement à \cite{mannilaDiscoveryFrequentEpisodes1997} qui recherche des ensembles fréquents et recherche ensuite des règles sur ces ensembles. De plus, l'architecture commune à RuleGrowth, TRuleGrowth et ERMiner permet de limiter la taille des règles recherchées. Il est également possible de limiter les éléments sur lesquels les règles sont recherchées. Dans notre cas d'utilisation, nous recherchons des règles dont les prédictions ne sont faites qu'à partir d'actionneurs. Ces algorithmes permettent cette limitation directement dans la recherche, ce qui réduit le temps total de calcul.\\
Cependant, ils ont un problème majeur dans le cas d'utilisation visé : ils prennent des transactions en entrée, au lieu d'une série temporelle. La notion de support dépend directement de la structure des transactions, et ne peut être appliquée en tant que telle sur une série temporelle. Ainsi, malgré les avantages de ces algorithmes, ils ne peuvent être appliqués en tant que tels à nos données d'entrée.
\subsection{Problèmes scientifiques}
A notre connaissance, les algorithmes de l'état de l'art ne sont pas assez satisfaisants pour résoudre le problème initial. Deux problèmes majeurs doivent être résolus :
\begin{enumerate}
	\item Comment définir le support d'une règle dans une série temporelle qui évite le problème de la section \ref{search-rulemining}?
	\item Comment construire un algorithme de fouille de règles sur cette nouvelle mesure de support ?
\end{enumerate}
De plus, cet algorithme doit aborder les points suivants :
\begin{enumerate}
	\setcounter{enumi}{2}
	\item Comment limiter la durée des règles trouvées ?
	\item Comment limiter la recherche à certains éléments dans la condition ou de la prédiction ?
	\item Comment éviter qu'une règle soit trouvée deux fois ?
\end{enumerate}
RuleGrowth répond aux points 4 et 5, mais ne prend que des transactions en entrée. TRuleGrowth, utilise une fenêtre glissante qui peut être utilisée pour répondre au problème 3 avec quelques modifications. Notre système utilise les principes de TRuleGrowth, mais les applique aux séries temporelles, pour traiter les deux premiers problèmes. La section suivante explique ces principes en détail.
\begin{figure}
	\centering
	\includegraphics[width=\linewidth]{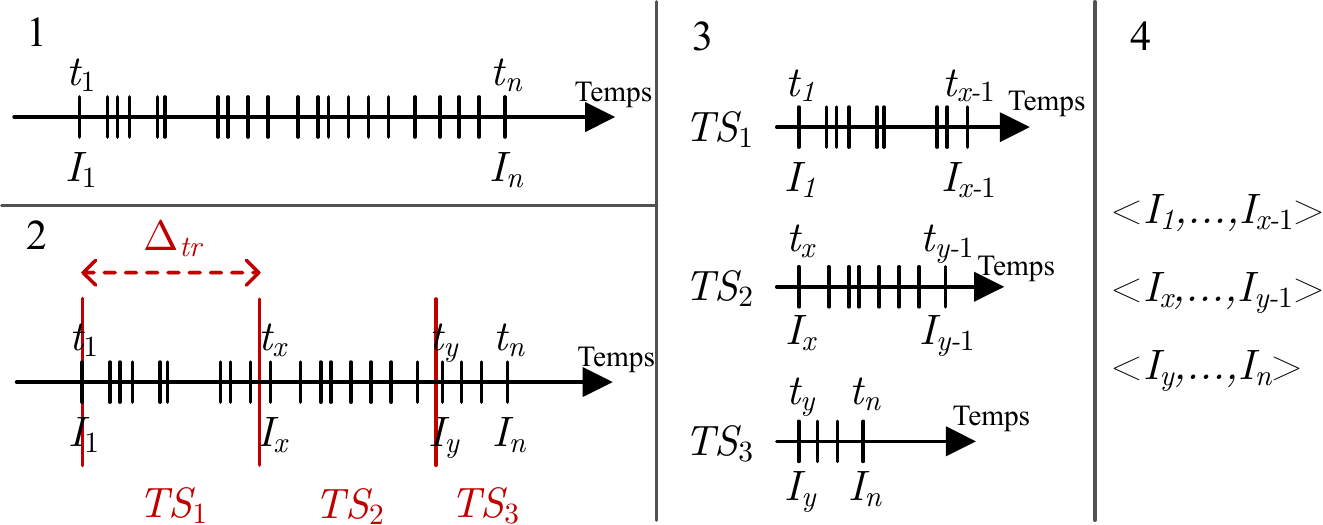}
	\caption{Exemple de conversion d'une série temporelle en transactions}
	\label{fig:TSConversion}
\end{figure}
\section{TRuleGrowth}\label{search-trulegrowth}
\subsection{Principes}\label{search-trulegrowthprinciples}
TRuleGrowth est un algorithme de recherche de règles semi-séquentielles sur des transactions. Il prend en entrée :
\begin{itemize}
	\item $TR = \{tr_1, ..., tr_{nb_{tr}}\}$: Un ensemble de transactions
	\item $min_{sup}$: La valeur minimale du support pour qu'une règle soit considérée comme fréquente
	\item $min_{int}$: La valeur minimale de l'intérêt pour qu'une règle soit acceptée comme règle de prédiction
	\item $window$: Un nombre maximal d'itemsets consécutifs dans lesquels les règles doivent se produire
\end{itemize}
Cet algorithme produit des règles semi-ordonnées, dont la condition et la prédiction sont des ensembles, comme indiqué dans la section \ref{search-motivation}.
Le support permet de vérifier qu'une règle est fréquente. Il existe deux types de support : absolu et relatif. Le support absolu $sup$ d'une règle est le nombre de transactions contenant cette règle. Il en va de même pour les éléments et les ensembles. Le support relatif $relSup$ d'un élément, règle ou ensemble est son support absolu divisé par le nombre total de transactions. Puis, l'intérêt d'une règle, via la mesure de confiance $conf$ \cite{azevedoComparingRuleMeasures2007}, permet vérifier sa fiabilité. Pour une règle $R: E_c \Rightarrow E_p$,
\begin{equation}
conf(R: E_c \Rightarrow E_p) =\frac{relSup(E_c \Rightarrow E_p)}{relSup(E_c)} 
\end{equation}
TRuleGrowth recherche les règles de manière incrémentale. Il cherche d'abord des règles basiques, dont la condition et la prédiction sont composées d'un seul élément. Puis, il essaie de les étendre progressivement, en y ajoutant un élément. TRuleGrowth est composé de trois sous-algorithmes : l'algorithme principal, ExpandLeft et ExpandRight. L'algorithme principal recherche les règles basiques dans la fenêtre, dont la condition et la prédiction sont composées d'un seul élément.
Si une règle a un support supérieur à $min_{sup}$, l'algorithme principal va essayer de l'étendre dans sa partie condition via ExpandLeft, et dans sa partie prédiction, via ExpandRight.\\
ExpandLeft et ExpandRight tentent d'étendre la règle en y ajoutant un élément, puis en calculant le support de la nouvelle règle formée. Si ce support est supérieur à $min_{sup}$, ExpandLeft et ExpandRight s'appelleront mutuellement, par récursivité, pour essayer de développer à nouveau la règle en ajoutant un élément. Pour éviter de chercher dans toutes les transactions, l'algorithme enregistre les transactions où la règle est apparue et recherche uniquement dans celles-ci. Ensuite, pour toutes les autres règles, leurs valeurs d'intérêt sont calculées pour les valider ou non.\\
Avec cette architecture, il est possible de trouver des doublons, c'est-à-dire la même règle plusieurs fois. TRuleGrowth évite de les trouver, grâce à deux mécanismes expliqués dans \cite{fournier-vigerMiningPartiallyOrderedSequential2015}. Premièrement, après une expansion de la condition, il n'est plus possible d'étendre la prédiction. Ainsi, ExpandLeft ne peut être suivi par ExpandRight, mais ExpandRight peut être suivi par ExpandLeft. Deuxièmement, une expansion n'est faite que sur des éléments plus grands selon l'ordre lexicographique. Par exemple, pour $\{b, c\} \Rightarrow \{x, y\}$, $\{b, c\}$ peut être étendu en ajoutant $d$, ou $e$, mais pas $a$, car il est inférieur à $c$ selon cet ordre.\\
Une idée pour appliquer TRuleGrowth à notre problème serait d'adapter les données en entrée, afin qu'elles puissent être acceptées par l'algorithme. Voici une proposition de modification et les inconvénients qui en découlent.
\subsection{Adaptation de la série temporelle}\label{search-trulegrowthadjustlent}
Pour résoudre le problème de structure d'entrée de ces algorithmes, on peut simplement convertir la séries temporelles en transactions, comme dans la figure \ref{fig:TSConversion}. Pour cela, la série temporelle est divisée (1 dans la figure) en séries plus petites d'une durée notée $\Delta_{tr}$ (2 et 3 dans la figure). Ensuite la notion de temps des petites séries est supprimée, pour ne garder que l'ordre d'apparition des éléments (4 dans la figure). Sans cette notion de temps, ce sont des séquences d'éléments, des transactions. Mais le principal problème de cette implémentation est le calcul du support d'une règle. Prenons un exemple avec trois transactions :
\begin{center}
	$\langle\{x\},\{x\},\{y\},\{x\},\{x\}\rangle$\\
	$\langle\{x\},\{y\},\{x\},\{x\},\{x\}\rangle$\\
	$\langle\{x\},\{x\},\{y\},\{x\},\{x\}\rangle$\\
\end{center}
Ici, $x \Rightarrow y$ est jugée valide, car son support est de 3, le même que $x$ et $y$. Si une règle n'a été vue qu'une fois dans une transaction, elle est jugée valide tout au long de cette transaction, même si elle aurait pu être invalidée, comme dans l'exemple : $x$ peut être vu sans $y$ après dans toutes les transactions. Le découpage d'une série en transactions peut conduire à des règles qui sont validées par erreur. Il y a d'autres problèmes, inhérents à $\Delta_{tr}$. Avoir un petit $\Delta_{tr}$ augmente le risque qu'une règle soit ``coupée en deux'', c'est-à-dire dont l'occurrence est séparée entre deux transactions, ce qui réduit l'intérêt. Avoir un gros $\Delta_{tr}$, sur une série temporelle, peut réduire le support absolu des règles recherchées par le système. La conversion d'une série temporelle en transactions peut être appliquée dans le cas d'utilisation proposé. Cependant, les limitations ci-dessus nous ont conduit à créer un nouvel algorithme, inspiré de TRuleGrowth, qui est pleinement adapté aux séries temporelles.
\section{TSRuleGrowth}\label{search-proposedalgorithm}
\subsection{Entrées, Sorties}\label{search-proposedalgorithm-inputsoutputs}
Ce nouvel algorithme recherche de règles de prédiction à partir d'une série temporelle d'éléments discrets. Cet algorithme est incrémental, et permet de limiter la recherche de règles à certains éléments dans la condition et de la prédiction. TSRuleGrowth prend en entrée :
\begin{itemize}
	\item $TS = \langle(t_1, I_1), ..., (t_{n}, I_{n})\rangle, I_1, ..., I_{n} \subseteq E$: Une série temporelle d'éléments discrets
	\item $min_{sup}$: La valeur minimale du support
	\item $min_{int}$: La valeur minimale de l'intérêt
	\item $window$: Une durée dans laquelle les règles doivent se produire
\end{itemize}
TSRuleGrowth produit des règles de prédiction semi-séquentielles utilisant des multiensembles, détaillées dans la section \ref{search-motivation-output}. Dans le cas d'utilisation proposé, la prédiction des règles ne contient que des éléments d'actionneurs.
\subsection{Métriques}\label{search-proposedalgorithm-metrics}
\subsubsection{Support}\label{search-support}
\begin{figure}[h]
	\centering
	\includegraphics[width=\linewidth]{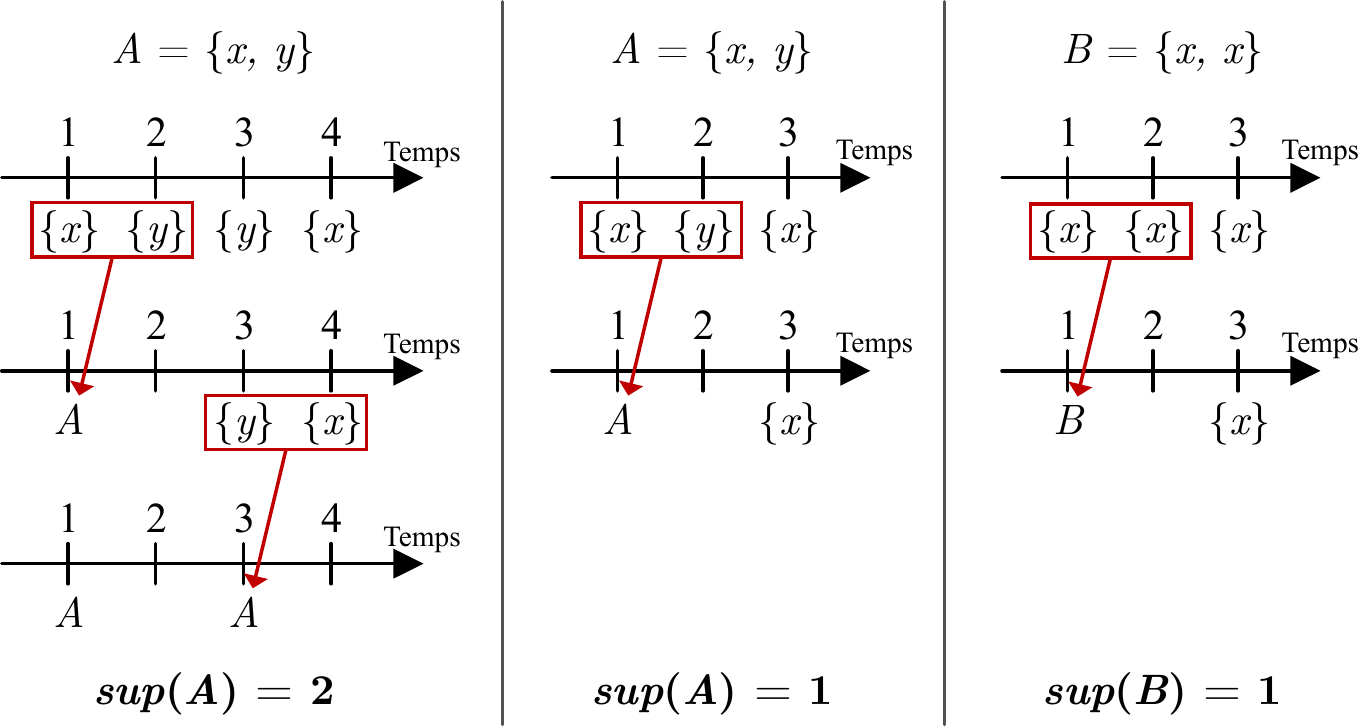}
	\caption{Exemples de calcul de support}
	\label{fig:SupportExamples}
\end{figure}
Pour une série temporelle $TS$ notée $\langle(t_1, I_1), ...., (t_{n_s}, I_{n_s})\rangle$ où $I_i$ est un itemset et $t_i$ un timestamp associé, le support de $x$, noté $sup(x)$, est défini comme le nombre d'itemsets contenant $x$.
\begin{equation}
sup(x) = \big\lvert(t_z, I_z) \in TS | x \in I_z \big\rvert
\end{equation}
Le support absolu d'un ensemble d'éléments $A$ est le nombre d'occurrences distinctes de tous les éléments de $A$ dans la fenêtre temporelle. Si une occurrence d'un élément de $A$ a contribué à une occurrence du multiensemble $A$, elle ne peut plus contribuer aux autres occurrences de $A$. Les exemples de la figure \ref{fig:SupportExamples} peuvent aider à comprendre ce concept plus facilement.
L'algorithme de comptage de support, Count, fait glisser une fenêtre sur la série temporelle. Si tous les éléments de $A$ sont vus, leurs occurrences sont mises sur liste noire, autrement dit blacklistés, pour les empêcher d'être impliqués dans une autre occurrence de $A$. Ceci permet de respecter la définition du support. Si plusieurs occurrences du même élément $A$ sont vues dans la même fenêtre, seules les plus anciennes seront blacklistées. Ceci laisse aux plus récentes, via une itération ultérieure, la possibilité de contribuer à une possible future occurrence de $A$. Le support absolu d'une règle $R : E_{c} \Rightarrow E_{p}$ est le nombre d'occurrences distinctes où tous les éléments de $E_{c}$ sont observés, suivi par tous les éléments de $E_{p}$. Les éléments de $E_{c}$ et $E_{p}$ ont aussi des listes noires, regroupées en deux ensembles, pour la condition et la prédiction.\\
Le support relatif d'un élément $x$, d'un multiensemble $A$ ou d'une règle $R$, noté $relSup$, est son support absolu divisé par le nombre total d'itemsets de la série temporelle.
\begin{equation}
relSup(R) = \frac{sup(R)}{|(t_z, I_z) \in TS|}
\end{equation}
Cette notion de support peut donc s'appliquer à des règles semi-ordonnées, contrairement à \cite{dasRuleDiscoveryTime1998, schluterAnalysisTimeSeries2011}, et évite le cas exprimé à la section \ref{search-rulemining}.
\begin{algorithm}
	\parbox{\linewidth}{
		\KwData{$A$: multiensemble, $TS = \langle(t_1, I_1), ..., (t_{n}, I_{n})\rangle,$ $I_1, ..., I_{n} \subseteq E$: série temporelle, $window$: durée}
		\tcp{Initialisation}
		Assigner liste noire $b(a)$ à chaque élément unique $a \in A$\;
		$sup(A) \leftarrow 0$\tcp*{Support de $A$}
		\tcp{Fenêtre glissante sur la série temporelle}
		\While{la fenêtre n'a pas atteint la fin de $TS$}{
			\textit{dist} $\leftarrow$ vrai\;
			Scanner la fenêtre, enregistrer les timestamps de $a \in A$ dans $T(a)$\;
			\ForEach{élément $a \in A$}{
				$T(a) \leftarrow T(a) \setminus b(a)$\;
				\If{$|T(a)| <$ multiplicité de $a$ dans $A$}{
					\textit{dist} $\leftarrow$ faux \tcp*{Pas d'occ. distincte}
				}
			}
			\If{dist est vrai}{
				$sup(A)$ += 1\;
				\ForEach{élément $a \in A$}{
					\tcp{Ajouter les plus anciens timestamps $T(a)$ à la liste noire de $a$}
					$m \leftarrow$ multiplicité de $a$ dans $A$\;
					$b(a) \leftarrow b(a) \cup m$ plus anciens timestamps de $T(a)$\;
				}
			}
			Itérer la fenêtre d'un itemset\;
		}
		\textbf{Renvoyer} $sup(A)$\;
	}
	\caption{Count}
\end{algorithm}
\subsubsection{Intérêt}\label{search-interest}
Dans TSRuleGrowth, on peut calculer l'intérêt d'une règle par sa confiance, conviction ou lift comme dit dans la section \ref{search-semisequential}. Dans le cas d'utilisation proposé, nous avons choisi netconf \cite{febrer-hernandezSPaCNFClassifierBased2016, ahnEfficientMiningFrequent2004}. Pour une règle $R: E_c \Rightarrow E_p$:
\begin{equation}
netconf(R) = \frac{relSup(R) - relSup(E_c) \times relSup(E_p)}{relSup(E_c)\times(1 - relSup(E_c))}
\end{equation}
Contrairement à la confiance, netconf teste l'indépendance entre les occurrences de $E_c$ et celles de $E_p$ \cite{ahnEfficientMiningFrequent2004}. Aussi, il est délimité entre -1 et 1, contrairement à conviction et lift, 1 montrant que $E_p$ a une forte probabilité d'apparaître après $E_c$, -1 que $E_p$ a une forte probabilité de ne pas apparaître après $E_c$, et 0 que cette probabilité est inconnue.
\subsection{Enregistrement d'occurrences de règles}\label{search-recording}
\begin{figure}[h]
	\centering
	\begin{minipage}[t]{.50\linewidth}	
		\centering
		\includegraphics[width=\textwidth]{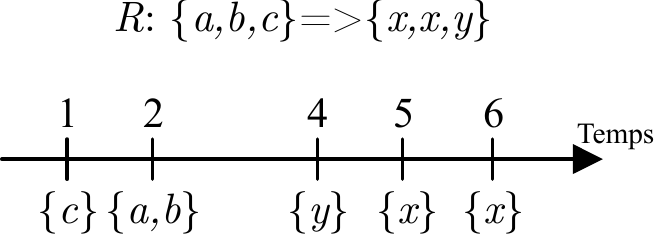}
		\caption{Exemple de règle et de série temporelle}
		\label{fig:occurrencerule}
	\end{minipage}
	\hfill
	\begin{minipage}[t]{.40\linewidth}
		\centering
		\includegraphics[width=.7\textwidth]{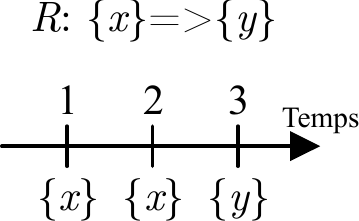}
		\caption{Exemple de règle et de série temporelle}
		\label{fig:newprinciples}
	\end{minipage}
\end{figure}

Prenons l'exemple d'une règle $R:\{a, b, c\} \Rightarrow \{x, x, y\}$.
Une occurrence de $R$ est décomposée comme l'occurrence de sa condition et de sa prédiction. En effet, un élément peut se trouver à la fois dans la condition et dans la prédiction, et il est nécessaire de distinguer les occurrences de cet élément dans la condition de celles dans la prédiction. L'occurrence d'un multiensemble est enregistrée dans un tableau associatif où les clés sont les éléments distincts du multiensemble, et leurs valeurs associées sont l'ensemble des timestamps où ces éléments sont observés. Dans la figure \ref{fig:occurrencerule}, l'occurrence de la condition de $R$ est$\{$$a$:$\{2\}$, $b$:$\{2\}$, $c$:$\{1\}$$\}$ et l'occurrence de la prévision est $\{$$x$:$\{5, 6\}$, $y$:$\{4\}$$\}$. Ici, deux timestamps sont enregistrés pour $x$, car il est présent deux fois dans la prédiction de $R$. Les occurrences d'un multiensemble sont stockés dans une liste de ces tableaux associatifs. Les occurrences de une règle sont enregistrées dans deux listes, pour la condition et la prédiction.
\subsection{Principes}\label{search-principles}
\subsubsection{Principes partagés avec TRuleGrowth}\label{search-sharedprinciples}
TSRuleGrowth reprend les principes de TRuleGrowth et les applique aux séries temporelles. Il utilise une fenêtre glissante, mais, contrairement à TRuleGrowth où la fenêtre est un nombre d'itemsets consécutifs, TSRuleGrowth a une fenêtre temporelle. Elle permet de restreindre la recherche, et d'avoir une estimation de la durée d'une règle. En outre, cet algorithme trouve des règles de base, où un élément en prédit un autre. Ensuite, récursivement, il les étendra, en ajoutant un élément dans la condition ou la prédiction, via ExpandCondition et ExpandPrediction. Ce mécanisme permet de limiter la longueur des règles à chercher, c'est-à-dire le nombre maximal d'éléments dans la condition et la prédiction.\\
Ensuite, TSRuleGrowth applique les deux principes de TRuleGrowth mentionnés dans la section \ref{search-trulegrowthprinciples}, pour éviter de trouver des règles doublons. Premièrement, ExpandPrediction ne peut pas être appelé par ExpandCondition. De plus, ExpandCondition et ExpandPrediction ne peuvent ajouter un élément que s'il est plus grand que les éléments de la condition ou de la prédiction, selon l'ordre lexicographique. Puisque TSRuleGrowth utilise une série temporelle comme entrée au lieu d'une liste de transactions, certaines notions doivent être redéfinies : le support, l'intérêt, et le stockage des occurrences d'une règle trouvée.
\subsubsection{Nouveaux principes}\label{search-newprinciples}
Prenons l'exemple illustré dans la figure \ref{fig:newprinciples}.
Pour cette règle $R$, même si $sup(R) = 1$, deux occurrences sont possibles : $\{$$x$:$\{1\}$, $y$:$\{3\}$$\}$ et $\{$$x$:$\{2\}$, $y$:$\{3\}$$\}$. Ce problème est inhérent aux séries temporelles : nous ne pouvons pas savoir \textit{a priori} quelle occurrence sera utile pour étendre cette règle. Pour ce faire, TSRuleGrowth essaie d'étendre toutes les occurrences vues de cette règle.\\
En outre, TSRuleGrowth n'utilise pas la même structure de règles que TRuleGrowth. Au lieu d'être des ensembles, la condition et la prédiction d'une règle sont des multiensembles, où les éléments peuvent apparaître plusieurs fois. Par conséquent, un principe issu de TRuleGrowth doit être modifié : ExpandCondition et ExpandPrediction peuvent ajouter un élément s'il est plus grand que les éléments de la condition ou prédiction, mais aussi s'il est égal au plus grand élément de ceux-ci, selon l'ordre lexicographique.\\
Mais un nouveau problème de doublons se pose. Prenons l'exemple de la figure \ref{fig:newprinciples}. Deux occurrences de la règle $\{x\}$$ \Rightarrow $$\{y\}$ sont observés : $\{$$x$:$\{1\}$, $y$:$\{3\}$$\}$ et $\{$$x$:$\{2\}$, $y$:$\{3\}$$\}$. Si nous étendons cette règle vers la règle $\{x, x\} \Rightarrow \{y\}$, la même occurrence sera trouvée deux fois. $\{$$x$:$\{1\}$, $y$:$\{3\}$$\}$ sera étendue à $\{$$x$:$\{1,\bm{2}\}$, $y$:$\{3\}$$\}$, en ajoutant le timestamp 2, et $\{$$x$:$\{2\}$, $y$:$\{3\}$$\}$ sera étendue à $\{$$x$:$\{\bm{1}, 2\}$, $y$:$\{3\}$$\}$, en ajoutant le timestamp 1. Pour éviter cette situation, et donc éviter la duplication, TSRuleGrowth fait la chose suivante : si la règle s'étend au plus grand élément de la condition ou de la prédiction, elle ne doit enregistrer que les timestamps de cet élément qui apparaissent \textbf{strictement plus tard} que le timestamp de ce même élément dans la règle de base. Ainsi, dans l'exemple précédent, la première occurrence est enregistrée, et non la seconde.
\subsection{Algorithme}\label{search-algorithm}
\subsubsection{Boucle principale}\label{search-mainloop}
Comme TRuleGrowth, la boucle principale de TSRuleGrowth tente de trouver des règles de base, c'est-à-dire des règles dont les conditions et les prédictions sont composées d'un seul élément. Pour ce faire, il calcule le support de toutes les règles basiques qui peuvent être créées dans la série temporelle. Si l'une de ces règles a un support supérieur à $min_{sup}$, elle essaie d'abord de l'étendre, en ajoutant un élément dans la condition (ExpandCondition), et dans la prédiction (ExpandPrediction). Enfin, elle calcule l'intérêt de cette règle pour la valider. Comme mentionné précédemment, l'algorithme recherche toutes les occurrences distinctes de la règle pour son support, mais aussi toutes les occurrences vues pour étendre la règle. Pour ce faire, TSRuleGrowth utilise un système de liste noire pour distinguer les occurrences.
\subsubsection{Extension des règles}\label{search-growingrules}
\begin{figure}
	\centering
	\begin{subfigure}[b]{\linewidth}
		\centering
		\includegraphics[width=.45\linewidth]{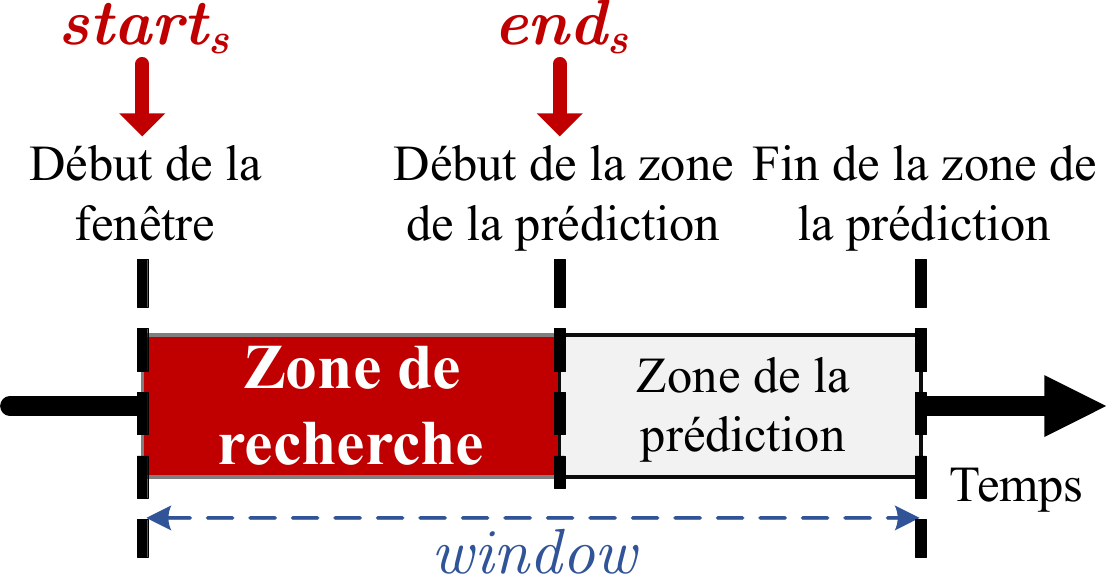}
		\captionof{figure}{Zone de recherche de ExpandCondition}
		\label{fig:ExpandConditionSearch}
	\end{subfigure}
	
	\begin{subfigure}[b]{\linewidth}
		\centering
		\includegraphics[width=.45\linewidth]{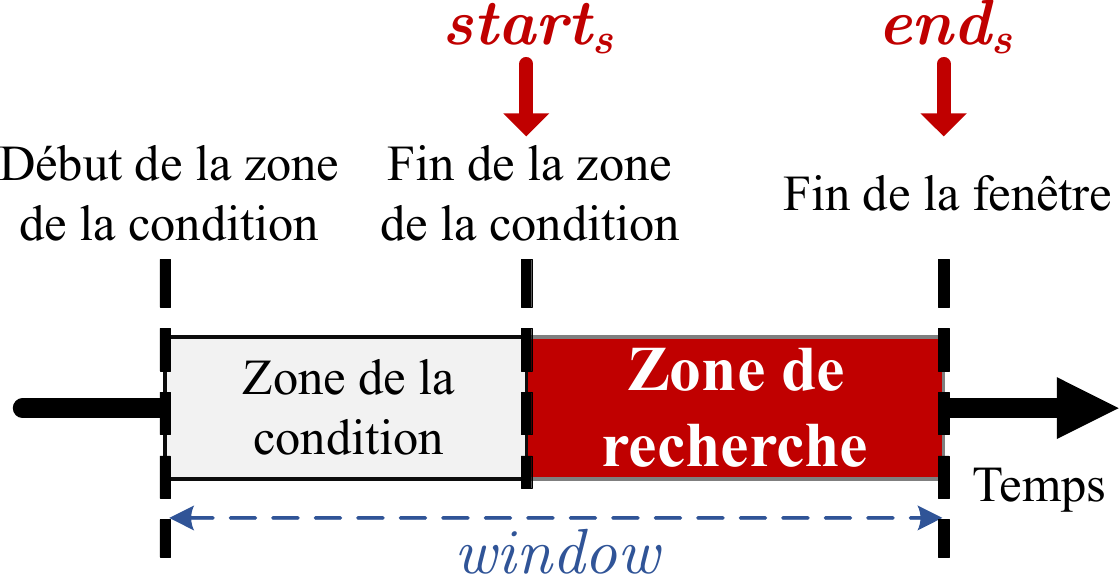}
		\captionof{figure}{Zone de recherche de ExpandPrediction}
		\label{fig:ExpandPredictionSearch}
	\end{subfigure}
	\caption{Zone de recherche pour étendre une règle}
\end{figure}
ExpandCondition essaie d'étendre une règle en ajoutant un élément à sa condition. Il passe en revue toutes les occurrences possibles de la règle, de la plus ancienne à la plus récente. Pour respecter la contrainte de temps imposée par $window$, la condition d'une règle ne peut s'étendre qu'entre deux timestamps, noté $start_s$ et $end_s$, montrés sur la figure \ref{fig:ExpandConditionSearch} : entre le début de la fenêtre, et le début de la prédiction. Comme pour ExpandCondition, ExpandPrediction cherche de nouveaux éléments pour la partie prédiction de la règle, à partir de la fin de la condition, et en respectant la taille de $window$ (figure \ref{fig:ExpandPredictionSearch}). Après avoir trouvé de nouvelles règles, ExpandCondition et ExpandPrediction essaient de les étendre à nouveau, et vérifient leur intérêt. Ici, les pseudocodes simplifiés de TSRuleGrowth et ExpandPrediction sont décrits.
\begin{algorithm}
	\parbox{\linewidth}{
		\KwData{$TS$: série temporelle, $min_{sup}$: support minimal, $min_{int}$: intérêt minimal, $window$: durée}
		Scanner $TS$ une fois. Pour chaque élément $e$, stocker les timestamps des itemsets contenant $e$ dans $T(e)$\;
		\tcp{Création de règles basiques}
		\ForEach{paire d'éléments i, j}{
			$sup(i \Rightarrow j) \leftarrow 0$\tcp*{Support de la règle}
			$O_c(i \Rightarrow j), O_p(i \Rightarrow j) \leftarrow [] $\tcp*{Occurrences}
			$b(i),b(j) \leftarrow \emptyset$\tcp*{Listes noires}
			\ForEach{$t_i \in T(i)$}{
				\ForEach{$t_j \in T(j)$}{
					\If{$0 < t_j - t_i \leq window$}{
						\tcp{Nouvelle occurrence}
						Ajouter $t_i$ à $O_c(i\Rightarrow j)$\;
						Ajouter $t_j$ à $O_p(i\Rightarrow j)$\;
						\If{$t_i \notin b(i)$ et $t_j \notin b(j)$}{
							\tcp{Nouvelle occurrence distincte}
							$sup(i \Rightarrow j)$ += 1\;
							$b(i) \leftarrow b(i) \cup \{t_i\}$\;
							$b(j) \leftarrow b(j) \cup \{t_j\}$\;
						}
					}
				}
			}
			\tcp{Expansion des règles basiques}
			\If{$sup(i \Rightarrow j)\geq min_{sup}$}{
				Lancer ExpandCondition et ExpandPrediction\;
				\lIf{netconf($\frac{|T(i)|}{|TS|}, \frac{|T(j)|}{|TS|},\frac{sup(i \Rightarrow j)}{|TS|} )$)$\geq min_{sup}$}{Afficher la règle}
			}
			
		}
	}
	\caption{TSRuleGrowth}
\end{algorithm}
\begin{algorithm}
	\parbox{\linewidth}{
		\KwData{$TS$: série temporelle, $E_{c} \Rightarrow E_{p}$: règle, $sup(E_{c})$, occurrences de $E_{c} \Rightarrow E_{p}$, $min_{sup}$: support minimal, $min_{int}$: intérêt minimal, $window$: durée}
		\tcp{Expansion de la règle basique $E_{c} \Rightarrow E_{p}$}
		\For{chaque occurrence de la règle $E_{c} \Rightarrow E_{p}$}{
			\ForEach{élément $k$ vu dans la zone de recherche}{
				\uIf{$k$ n'a jamais été vu avant}{
					Créer une règle $E_{c} \Rightarrow E_{pk}$, sa liste d'occurrences et ses listes noires\;
					$sup(E_{c} \Rightarrow E_{pk})$ $\leftarrow$ 0\;
				}
				\ForEach{timestamp de $k$ $t_k$ dans la fenêtre (ordre croissant)}{
					\If{$k > \max(e), e \in E_{p}$ où $t_k >$ occurrences de $k$ dans la partie prédiction de la règle}{
						Créer nouvelle occurrence de $E_{c} \Rightarrow E_{pk}$\;
						\If{timestamps pas dans listes noires}{
							$sup(E_{c} \Rightarrow E_{pk})$ += 1\;
							Ajouter timestamps aux listes noires\;
						}
					}
				}
			}
		}
		\tcp{Expansion des nouvelles règles trouvées}
		\ForEach{k où $sup(E_{c} \Rightarrow E_{pk}) \geq min_{sup}$}{
			$sup(E_{pk}) \leftarrow Count(E_{pk}, TS, window)$\;
			Lancer ExpandCondition et ExpandPrediction\;
			\lIf{netconf$(\frac{sup(E_{c})}{|TS|}, \frac{sup(E_{pk})}{|TS|}, \frac{sup(E_{c} \Rightarrow E_{pk})}{|TS|}) \geq min_{int}$}{Afficher la règle}
		}
	}
	\caption{ExpandPrediction}
\end{algorithm}
\section{Expérimentations et résultats}\label{search-results}
\begin{figure*}
	\centering
	\begin{minipage}[t]{.30\textwidth}
		\centering
		\includegraphics[width=\linewidth]{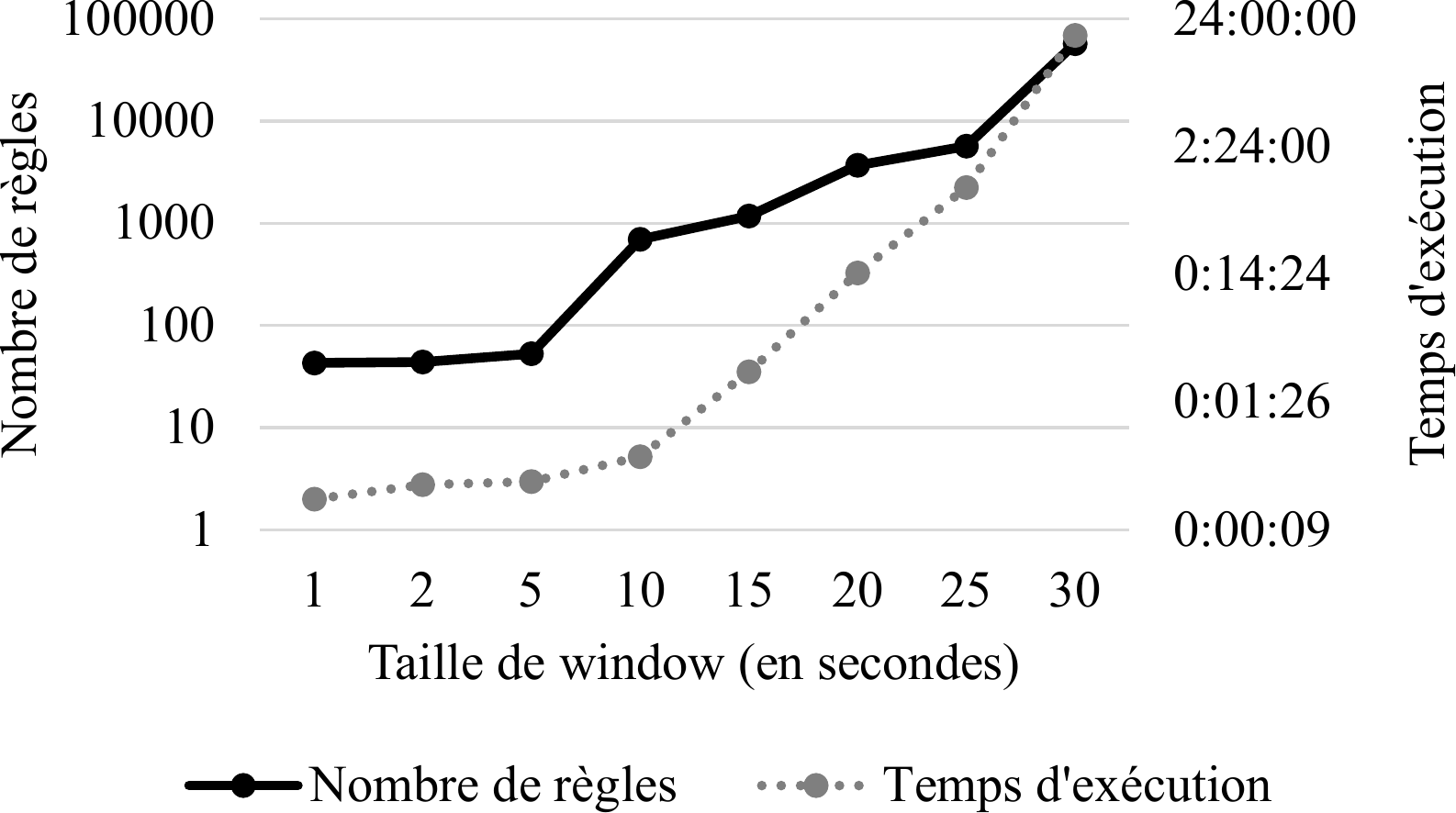}
		\captionof{figure}{Nombre de règles trouvées par TSRuleGrowth et temps d'exécution}
		\label{fig:RulesExecution}
	\end{minipage}
	\hfill
	\begin{minipage}[t]{.30\textwidth}
		\centering
		\includegraphics[width=\linewidth]{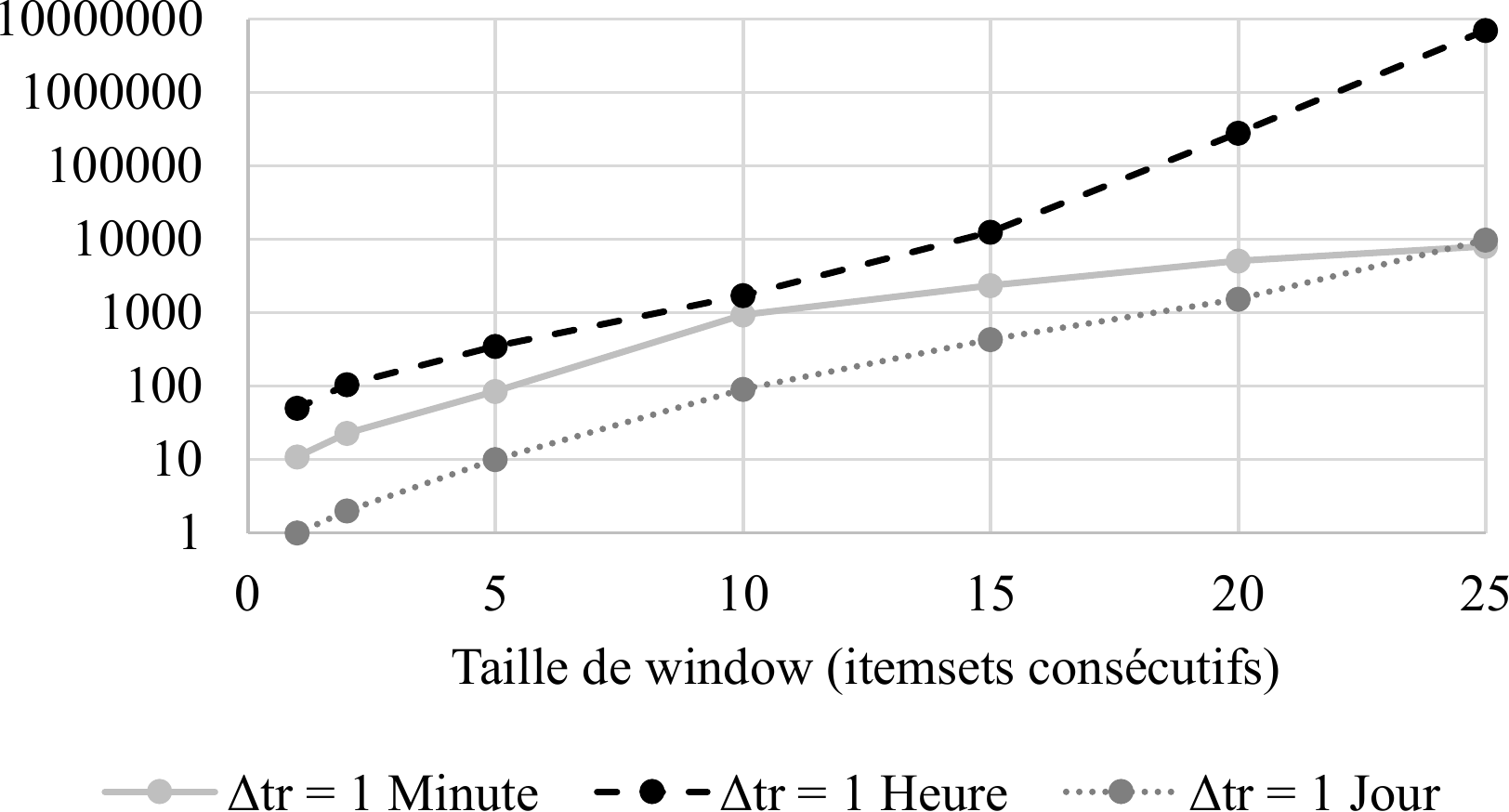}
		\captionof{figure}{Nombre de règles trouvées par TRuleGrowth}
		\label{fig:TotalRulesFoundTRG}
	\end{minipage}%
	\hfill
	\begin{minipage}[t]{.30\textwidth}
		\centering
		\includegraphics[width=\linewidth]{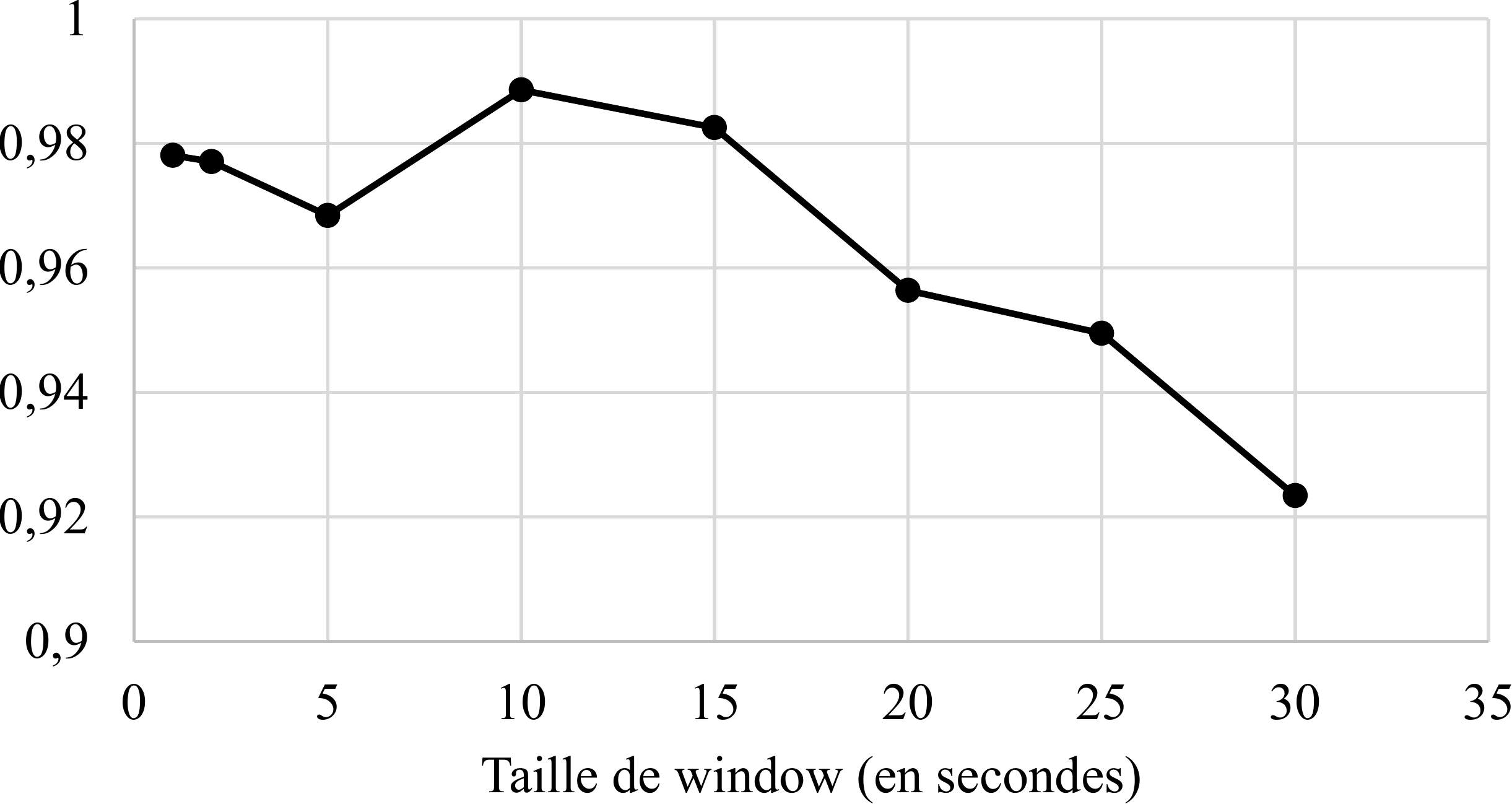}
		\captionof{figure}{Moyenne de l'intérêt des règles trouvées par TSRuleGrowth}
		\label{fig:MeanInterest}
	\end{minipage}%
\end{figure*}
Nous avons testé cet algorithme sur la base de données Orange4Home \cite{cuminDatasetRoutineDaily2017}, qui enregistre les activités quotidiennes d'un occupant. Elle contient 180 heures de données, sur une période de 4 semaines consécutives, à partir de 236 objets connectés intégrés dans un appartement. Pour les besoins de l'expérience, certains objets ont été spécifiés manuellement comme actionneurs : volets, portes et luminaires par exemple. De plus, un processus de discrétisation de l'amplitude a été effectué sur des objets qui rapportaient des données continues, comme un capteur de température. Pour rappel, seuls les actionneurs peuvent fournir des éléments pour la prédiction des règles. TSRuleGrowth a été implémenté en Python\footnote{Python 3.7.3, CPU: Intel(R) Xeon(R) Gold 5118 @ 2.30GHz, RAM: 128GiB, Ubuntu 18.04.2 LTS, Multiprocessing ajouté au code}, avec $min_{sup}=20$, $min_{int}=0.9$, et un $window$ de 1, 2, 5, 10, 15, 20, 25 et 30 secondes. TSRuleGrowth trouve des règles simples lorsque $window$ est petit. Les règles suivantes ont été trouvées par TSRuleGrowth dans une fenêtre de deux secondes :
\begin{itemize}
	\item \{'bedroom switch top right: ON'\} $\Rightarrow$ \{'bedroom light 1: 0', 'bedroom light 2: 0'\} : le bouton en haut à droite de la chambre éteint les lumières de cette pièce.
	\item \{'livingroom switch 2 top right: ON'\} $\Rightarrow$ \{'livingroom shutter 1: 100', 'livingroom shutter 2: 100', 'livingroom shutter 3: 100', 'livingroom shutter 4: 100', 'livingroom shutter 5: 100'\} : le bouton  en haut à droite du deuxième interrupteur du salon commande tous les volets.
\end{itemize}
Ces règles décrivent des prédictions à court-terme, telles que les actions des interrupteurs dans l'environnement. Ainsi, avec une petite fenêtre temporelle, l'algorithme peut déjà décrire certains des mécanismes de l'environnement connecté. Ensuite, lorsque la fenêtre est plus grande, des règles plus complexes sont découvertes en plus des règles simples, prenant en compte des objets plus diversifiés, pour caractériser des situations plus complexes. Ces règles, puisque la fenêtre d'observation est plus grande, peuvent révéler les habitudes de l'utilisateur. La règle \{'bathroom door: OPEN', 'kitchen presence: OFF', 'walkway light: 0'\} $\Rightarrow$ \{'bathroom light1: 100', 'bathroom light2: 100'\}, vue dans une fenêtre de 30 secondes, décrit la situation où l'occupant quitte la cuisine et entre dans la salle de bains. Il est à noter que le nombre de règles trouvées par l'algorithme augmente de façon exponentielle à mesure que la fenêtre grandit, comme le montre la figure \ref{fig:RulesExecution}. En effet, la plupart des règles trouvées sur une fenêtre temporelle seront trouvées sur une fenêtre plus grande, en plus des nouvelles règles. De plus, lorsque la fenêtre grandit, davantage d'objets peuvent être utilisés pour décrire une situation, et davantage de règles peuvent être validées en conséquence.\\
Regardons maintenant les résultats rapportés par TRuleGrowth, sur la figure \ref{fig:TotalRulesFoundTRG}. TRuleGrowth a été exécuté avec les mêmes paramètres que TSRuleGrowth. Deux variations ont été faites : la longueur $\Delta_{tr}$ des transactions, et la taille de $window$. De plus, la mesure d'intérêt utilisée est netconf. Selon la longueur $\Delta_{tr}$ assignée aux transactions lors du découpage de la série temporelle, le nombre de règles trouvées peut varier considérablement, comme expliqué dans la section \ref{search-trulegrowthadjustlent}. TSRuleGrowth se libère de cette limitation. Plus la fenêtre est grande, plus l'espace de recherche est grand. Par conséquent, le temps d'exécution de TSRuleGrowth augmente exponentiellement à mesure que la fenêtre augmente, comme le montre la figure \ref{fig:RulesExecution}. On trouve de plus en plus de règles pour décrire les situations. Ces situations, impliquant l'utilisateur, ne peuvent pas être aussi certaines que les règles simples vues avant, telles que celles d'un interrupteur. En conséquence, l'intérêt moyen des règles tend à diminuer au fur et à mesure que la fenêtre augmente, comme le montre la figure \ref{fig:MeanInterest}.
\section{Conclusion}\label{search-conclusion}
Cet article décrit deux contributions : une nouvelle notion de support sur une série temporelle, et un algorithme de recherche de règles de prédiction semi-ordonnées sur une série temporelle d'éléments discrets. La notion de support est libéré des limites exprimées dans l'état de l'art, et l'algorithme se distingue également par ses caractéristiques. Tout d'abord, une architecture incrémentale, inspirée de TRuleGrowth, permettant de limiter la recherche à certains éléments si nécessaire, comme dans le cas d'utilisation proposé. Une fenêtre glissante permet de limiter la durée des règles recherchées. Un nouveau mécanisme évite de trouver plusieurs fois la même règle. Les résultats présentés permettent de tester et valider l'algorithme sur des données réelles provenant d'un environnement connecté. Ils montrent des règles de prédiction simples, telles que l'action d'un interrupteur dans une pièce donnée, et d'autres plus complexes, impliquant des objets connectés différents. Ces dernières règles ouvrent sur des propositions d'automatisation pertinentes aux utilisateurs d'un système d'intelligence ambiante.
\bibliography{citations}
\end{document}